\documentclass[10pt,twocolumn,letterpaper]{article}

\usepackage{cvpr}
\usepackage{times}
\usepackage{epsfig}
\usepackage{graphicx}
\usepackage{amsmath}
\usepackage{amssymb}
\usepackage{amsfonts}
\usepackage{amsmath}
\usepackage{algorithm}
\usepackage{algorithmic}
\usepackage{color}
\usepackage{multirow}
\usepackage{threeparttable}

\usepackage[breaklinks=true,bookmarks=false]{hyperref}

\cvprfinalcopy 

\def\cvprPaperID{****} 

\newcommand{\ptsh}{PTS\textsuperscript{3}H}

\begin{document}


\title{Pairwise Teacher-Student Network for Semi-Supervised Hashing}

\author{Shifeng Zhang, Jianmin Li and Bo Zhang\\
Institute for Artificial Intelligence, State Key Lab of Intelligent Technology and Systems, \\
Beijing National Research Center for Information Science and Technology, \\
Department of Computer Science and Technology, Tsinghua University, Beijing 100084, China\\
{\tt\small zhangsf15@mails.tsinghua.edu.cn, lijianmin@mail.tsinghua.edu.cn, dcszb@mail.tsinghua.edu.cn}
}

\maketitle

\begin{abstract}
Hashing method maps similar high-dimensional data to binary hashcodes with smaller hamming distance, and it has received broad attention due to its low storage cost and fast retrieval speed. Pairwise similarity is easily obtained and widely used for retrieval, and most supervised hashing algorithms are carefully designed for the pairwise supervisions. As labeling all data pairs is difficult, semi-supervised hashing is proposed which aims at learning efficient codes with limited labeled pairs and abundant unlabeled ones. Existing methods build graphs to capture the structure of dataset, but they are not working well for complex data as the graph is built based on the data representations and determining the representations of complex data is difficult. In this paper, we propose a novel teacher-student semi-supervised hashing framework in which the student is trained with the pairwise information produced by the teacher network. The network follows the smoothness assumption, which achieves consistent distances for similar data pairs so that the retrieval results are similar for neighborhood queries. Experiments on large-scale datasets show that the proposed method reaches impressive gain over the supervised baselines and is superior to state-of-the-art semi-supervised hashing methods.
\end{abstract}

\section{Introduction}

With the explosion of high-dimensional media data, approximate Nearest Neighbor(ANN) search~\cite{gionis1999similarity} has attracted broad attention for efficient information retrieval. Among the existing ANN methods, hashing has become a popular tool for ANN search on large-scale datasets due to its fast search time and small storage space~\cite{gionis1999similarity,liu2012supervised,Shen_2015_CVPR,xia2014supervised,weiss2009spectral}. It aims at encoding high-dimensional data into compact hashcodes, so that similar data are mapped to hashcodes with similar hamming distance.

Among the existing hashing methods, data-dependent learning-to-hash methods aim at learning hash functions with the training data, and the learned codes is able to capture the data distributions. Learning-to-hash methods can be divided into three categories: unsupervised hashing~\cite{gong2013iterative,weiss2009spectral}, supervised hashing~\cite{liu2012supervised,Shen_2015_CVPR} and semi-supervised hashing~\cite{zhang2017ssdh,yan2017semi,qiu2017deep}. Experiments convey that the codes learned by (semi-)supervised hashing methods can capture more semantic information than unsupervised ones. Recently, with the rapid development of deep learning~\cite{krizhevsky2012imagenet,he2015deep}, deep hashing methods have achieved great success~\cite{xia2014supervised,lai2015simultaneous,liu2016deep,zhang2017scalable,zhang2017ssdh,li2017deep,cakir2018hashing}. It aims at learning hashcodes and the deep networks simultaneously, thus the codes generated by deep networks contain much better semantic information.

For ANN search, pairwise similarities between data pairs play an important role in evaluating the quality of search. For generating efficient hashcodes, (deep) supervised hashing problems regard the pairwise similarity as the basic supervision such that similar data pairs should be mapped to codes with small hamming distance. Most hashing methods model the similarities with the pairwise losses, and optimizing them is expected to generate the codes where the hamming distances are accordant with the similarities. For ease of back-propagation, these methods simply generate data pairs within a mini-batch and achieve good results~\cite{liu2016deep,cao2016deep,lu2017deep,li2015feature}.


Despite the success of supervised hashing, labeling all the database data (pairs) is almost intractable as the number of data is dramatically increasing. To utilize the abundant database data, deep semi-supervised hashing~\cite{zhang2017ssdh,yan2017semi} has been proposed in which the hash function is trained with the labeled data pairs and abundant unlabeled ones. The success of semi-supervised hashing lies in the smoothness assumption such that neighborhood data are likely to have the same predictions. These methods construct graphs for the unlabeled data to capture the neighborhood structure among the samples. However, the data and their representations may lie in high-dimensional nonlinear manifolds, especially for complex data like images and videos, and the representations may not learn well with limited data. As the graph is built based on data representations, the graph may not model the neighborhood structure of data precisely, which violates the smoothness assumption to some extent and affect the hashing performance.


Recently, perturbation-based teacher-student semi-supervised learning (SSL) algorithms have witnessed great success~\cite{laine2016temporal,tarvainen2017mean}. These methods follow the smoothness assumption in which the learned classifiers produce consensus prediction of a perturbed input, thus they can better capture the structure of unlabeled data~\cite{zhu2003semi} and produce better representations for the graph training~\cite{luo2017smooth}. However, the proposed teacher-student method can just deal with data with single label, but does not consider the pairwise relationship between samples, which is crucial for semi-supervised hashing. By carefully designing the teacher-student architecture and the loss for pairwise similarities, we may utilize the advantage of this architecture and obtain a novel semi-supervised hashing method.

In this paper, we propose a novel semi-supervised hashing algorithm called {\em Pairwise Teacher-Student Semi-Supervised Hashing}(\ptsh) in which the pairwise similarities are used for supervision and abundant unlabeled data pairs are provided. The proposed \ptsh{} is a teacher-student network architecture where the student is trained with pairwise loss and unsupervised regularizers, and the teacher is the average of the student network to generate efficient pairwise representations. As hashing mainly focuses on the pairwise information, we propose the general {\em consistent pairwise loss} such that similar queries produce similar pairwise similarities with the database and achieve similar retrieval results, aiming at following the smoothness assumptions~\cite{tarvainen2017mean,laine2016temporal}. For modeling pairwise similarities between samples with local and global pairwise information, we propose two types of losses: {\em consistent similarity loss} for consistent pairwise similarities among data, and {\em quantized similarity loss} in which the quantized~\cite{jegou2011product} similarities can be modeled by the teacher network by global data pairs. Experiment shows that the proposed \ptsh{} achieves great improvement over the supervised baselines, and it is superior or comparable with the state-of-the-art semi-supervised hashing algorithms.

\section{Background}

Suppose we are given $n$ data samples $\mathbf{x}_1, \mathbf{x}_2, ..., \mathbf{x}_n \in \mathcal{X}$, and $\mathcal{X}$ is the training dataset. Denote $\mathcal{S}$ as a set such that $(i,j) \in \mathcal{S}$ implies $\mathbf{x}_i, \mathbf{x}_j$ have similarity information, and we denote $s_{ij}=1$ if $\mathbf{x}_i, \mathbf{x}_j$ are similar, and $s_{ij}=0$ otherwise. In practical applications, the similarity information of some data pairs is unknown. We denote $\mathcal{U}$ as the pairs where the pairwise similarity information is unknown.

Denote $b$ as the length of the hashcode to learn, the goal of the semi-supervised hash learning is to learn the hash function $H(\mathbf{x}) = [h_1(\mathbf{x}), ..., h_b(\mathbf{x})]^\top \in \{-1,1 \}^b$ with $n$ data samples and the pairwise similarities. We denote $\mathbf{h}_i = H(\mathbf{x}_i), i = 1,2,...,n$ as the learned hashcode of $\mathbf{x}_i$.

\begin{figure*}[t]
    \setlength{\abovecaptionskip}{0pt}
    \setlength{\belowcaptionskip}{0pt}
    \centering
    \includegraphics[scale=0.7]{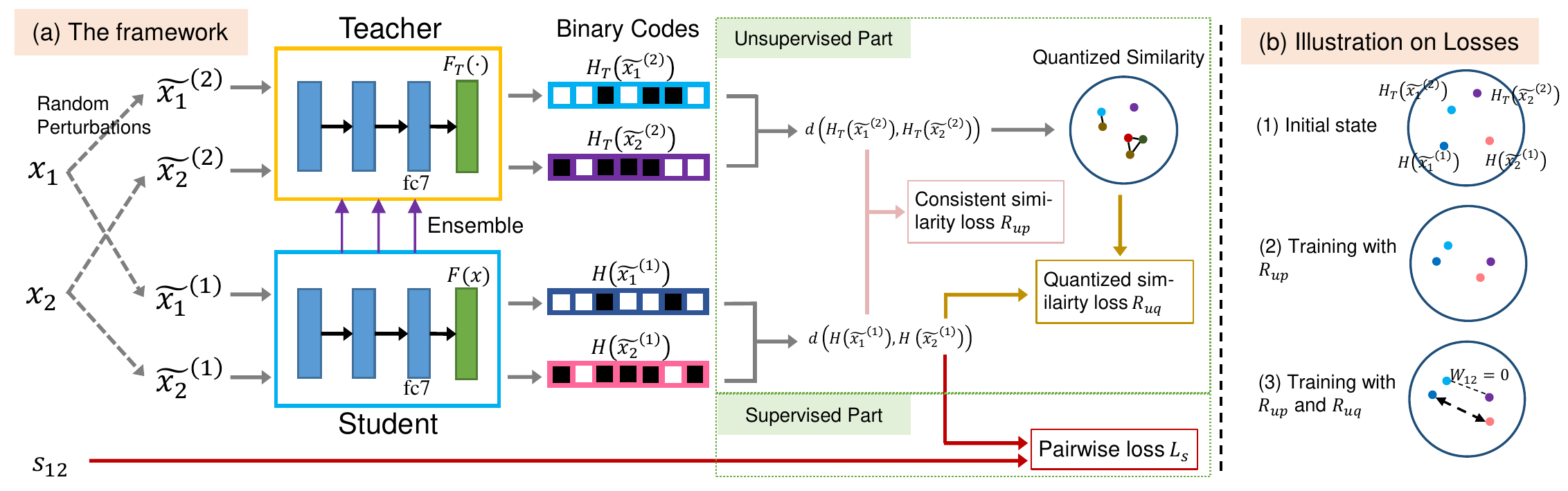}
    \caption{Overview of the \ptsh{} algorithm. (a) The general framework of the \ptsh{} algorithm. The input data is $\mathbf{x}_1, \mathbf{x}_2$ and the pairwise similarity is $s_{12}$ if available. $\mathcal{L}_s$ is computed for labeled pairs and $\mathcal{R}_{uc}, \mathcal{R}_{up}, \mathcal{R}_{uq}$ are computed for unlabeled pairs. (b) Illustration on training with different unsupervised regularizations.}
    \label{fig:framework}
\end{figure*}

\subsection{Pairwise Loss for Supervised Hashing}

Pairwise losses is widely used for solving (deep) supervised hashing algorithm with pairwise similarity as supervision~\cite{liu2012supervised,lin2014fast,liu2016deep,cao2016deep,li2015feature,yan2017semi}. For the given training data and pairwise information, the basic formulation of pairwise loss is 
\begin{equation}
\mathcal{L}_s = \frac{1}{|\mathcal{S}|} \sum_{(i,j) \in \mathcal{S}} l(u_{ij}, s_{ij}), \quad u_{ij} = \mathrm{sim}(\mathbf{h}_i, \mathbf{h}_j)
\label{eq:pls}
\end{equation}
where $u_{ij}=\mathrm{sim}(\mathbf{h}_i, \mathbf{h}_j)$ are the similarity (or distance) between the codes $\mathbf{h}_i, \mathbf{h}_j$.

Different types of $l(u_{ij}, s_{ij})$ are discovered in different supervised hashing algorithms such that
\begin{itemize}
\item KSH loss: $l(u_{ij}, s_{ij}) = [b(2s_{ij}-1) - u_{ij}]^2, u_{ij}=\mathbf{h}_i^\top \mathbf{h}_j$ in KSH~\cite{liu2012supervised} and FastH~\cite{lin2014fast};
\item DSH loss: $l(u_{ij}, s_{ij}) = -s_{ij} u_{ij} + (1-s_{ij}) \max(0, 2b+u_{ij}), u_{ij}=-(\mathbf{h}_i- \mathbf{h}_j)^2$ in DSH~\cite{liu2016deep};
\item DPSH loss: $l(u_{ij}, s_{ij}) = -s_{ij} u_{ij} + \log (1+e^{u_{ij}}), u_{ij} = \frac{1}{2} \mathbf{h}_i^\top \mathbf{h}_j$ in DPSH~\cite{li2015feature}, DHN~\cite{cao2016deep}.
\end{itemize}

Optimizing $l(u_{ij}, s_{ij})$ is expected to learn hashcodes such that similar data pairs have codes with small hamming distance, and vice versa. It should be noticed that the supervised information is just pairwise information, which is widespread in the real world.

\subsection{Semi-Supervised Hashing}

Semi-supervised hashing focuses on learning hash function with limited labeled data pairs as well as abundant unlabeled pairs in the database. Similar with semi-supervised learning, the general form of loss to be optimized is
\begin{equation}
\mathcal{L} = \mathcal{L}_s + \omega \mathcal{R}_u
\label{eq:ssh}
\end{equation}
where $\mathcal{L}_s$ is Eq. (\ref{eq:pls}), $\mathcal{R}_u$ ls the regularization term for unlabeled data. SPLH~\cite{wang2012semi} adopts the bit-balanced constraint for regularization, but it does not consider the relationship between samples. Graph-based methods like SSDH~\cite{zhang2016ssdh} and BGDH~\cite{yan2017semi} construct an affinity graph for indicating pairwise similarities between unlabeled samples, and the regularization loss is constructed based on the graph. These methods succeed in capturing the neighborhood structures between samples, but the graph is constructed by data representations, and the semantic gap may be involved among the representations, which may violate the smoothness assumptions. Recently, deep generative models have achieved success in semi-supervised learning problems, and DSH-GANs~\cite{qiu2017deep} proposes a GAN~\cite{salimans2016improved} based hashing method. The conditional GAN is trained with labeled and unlabeled data to generate labeled samples, which are used for training the hashing network. It achieves state-of-the-art in some datasets, it is not able to be trained with pairwise supervision.

\subsection{Teacher-Student Network for Semi-Supervised Learning}

Semi-supervised learning (SSL) aims at learning with limited labeled data and abundant unlabeled data. Most semi-supervised learning methods lies in the smoothness assumption such that similar data correspond to the same label. Various approaches are discovered such as transductive approach~\cite{krause2010discriminative,weston2012deep}, graph-based methods~\cite{belkin2006manifold,zhu2003semi}, but they are not working well in complex dataset as the underlying structure of data is hard to capture. Recently, perturbation-based semi-supervised learning approach has achieved great success, where a perturbed input corresponds to the consensus prediction. These methods propose a dual role, i.e., the teacher and the student. The student is learned as before; the teacher generates the targets for training the student. Formally, considering the dataset $\mathcal{X}$ where part of data are labeled, we aim at optimizing the following loss function:
\begin{equation}
\mathcal{L}^{(c)} = \mathcal{L}^{(c)}_{s} + \omega \mathcal{R}^{(c)}_{u}
\label{eq:ssl}
\end{equation}
where $c$ denotes \underline{c}lassification, $\mathcal{L}^{(c)}_{s}$ is the supervised term such as the softmax loss, $\mathcal{R}^{(c)}_{u}$ is the unsupervised regularization such that
\begin{eqnarray}
\mathcal{R}^{(c)}_{u} = \sum_{\mathbf{x} \in \mathcal{X}} d(f(\tilde{\mathbf{x}}^{(1)}), f_T(\tilde{\mathbf{x}}^{(2)}))
\label{eq:ssl_r}
\end{eqnarray}
where $\tilde{\mathbf{x}}^{(1)}, \tilde{\mathbf{x}}^{(2)}$ are two random perturbations, $f(\cdot), f_T(\cdot)$ are the outputs of student and teacher network respectively, and $d(\cdot, \cdot)$ is the distance between two features. There are several ways to define the teacher $f_T$. TempEns~\cite{laine2016temporal} considers $f_T$ as the exponentially moving average(EMA) of the student's output; Mean Teacher~\cite{tarvainen2017mean} averages the weights of student with EMA to form the teacher network; VAT~\cite{miyato2018virtual} introduces the adversarial perturbations instead of random perturbations. These methods achieve state-of-the-art on SSL problems.

In spite of this, perturbation-based methods is just able to regularize the single data point, but do not consider neighborhood structure between samples. SNTG~\cite{luo2017smooth} constructs a graph by the teacher to capture the neighborhood structure, and introduces a pairwise regularization term with the graph. Experiments convey that the additional term achieves better performance as both the consistency of the perturbed data and neighborhood samples is considered. However, the graph in SNTG is built specifically for classification.


With the success of teacher-student network for semi-supervised learning, in this paper, we propose a novel teacher-student framework for semi-supervised hashing in which only small portion of pairwise similarity information is provided. Considering we perform the hamming distance learning, we propose a novel {\em consistent pairwise loss} in which the consistent feature distances for similar data pairs are reached so that it is able to follow the smoothness assumption where neighborhood queries achieve similar retrieval results. Experiments show its superiority over the state-of-the-art semi-supervised hashing algorithms.

\section{Methodology}

In this section, we propose the novel deep semi-supervised hashing called {\em Pairwise Teacher-Student Semi-Supervised Hashing(\ptsh)}, in which the teacher-student network is adopted. 

\subsection{The Teacher-Student Framework}

The proposed \ptsh{} is a teacher-student architecture shown in Figure \ref{fig:framework}(a). The architecture of teacher network and the student are the same, in which the last layer is the fully-connected layer with $b$ outputs ($b$ is the hashcode length), and the rest layers can be the basic deep network like AlexNet, VGGNet, etc.

The update rule of the teacher-student network is similar as Mean Teacher~\cite{tarvainen2017mean}. The student is learned with labeled data pairs and guided by the teacher. Denote $\theta(t)$ and $\theta_T(t)$ as the parameters of the student and teacher network at training step $t$ respectively, the teacher network is updated by EMA as follows:
\begin{equation}
\theta_T(t) = \alpha \theta_T(t-1) + (1-\alpha) \theta(t)
\label{eq:t_ema}
\end{equation}
thus the teacher is the average embedding of the student, and the teacher's output can be regarded as the mean embedding of the student's.

Denote $F(\mathbf{x}), F_T(\mathbf{x}) \in \mathbb{R}^r$ as the output of the student and teacher networks respectively, the binary codes of data $\mathbf{x}$ can be easily obtained with either the student network such that $H(\mathbf{x}) = \mathrm{sgn}(F(\mathbf{x}))$, or the teacher network $H_T(\mathbf{x}) = \mathrm{sgn} (F_T(\mathbf{x}))$. Note that the $\mathbf{x}$ is not perturbed in the code generation.

\subsection{Loss Function}
\label{sec:loss}

The general form of loss to be optimized is Eq. (\ref{eq:ssh}). For labeled data pairs, the training loss is the pairwise loss function in Eq. (\ref{eq:pls}). For training with the unlabeled data, $\mathcal{R}_u$ should be defined in which the teacher network generates targets to guide the student network. As hash learning focuses on the pairwise similarities of the codes, learning the similarities of the embedded hamming space are quite important. For input pairs, the targets for the student should be the similarities of the codes generated by the teacher. We therefore propose the general form of the {\em consistent pairwise loss} such that
\begin{equation}
\begin{split}
\mathcal{R}_u &= \frac{1}{|\mathcal{X}|^2} \sum_{\mathbf{x}_1, \mathbf{x}_2 \in \mathcal{X}} l_c (u_{12}, u_{T12}) \\
u_{12} &=\mathrm{sim}(H(\tilde{\mathbf{x}_1}^{(1)}), H(\tilde{\mathbf{x}_2}^{(1)})) \\
u_{T12} &= \mathrm{sim}(H_T(\tilde{\mathbf{x}_1}^{(2)}), H_T(\tilde{\mathbf{x}_2}^{(2)}))
\end{split}
\label{eq:ssl_u}
\end{equation}
where $\tilde{\mathbf{x}_{i}}^{(1)}, \tilde{\mathbf{x}_{i}}^{(2)}, i=1,2$ are two random perturbations of $\mathbf{x}_{i}$, $l_c(u, u_T)$ is a certain type of loss and $u, u_T$ denote the pairwise similarities of codes generated from the student and the teacher respectively. Eq. (\ref{eq:ssl_u}) is quite different from the original Mean Teacher~\cite{tarvainen2017mean} in which only the single data point is considered for training.

For Eq. (\ref{eq:ssl_u}), We propose two simple but efficient form of losses named {\em consistent similarity loss} and {\em quantized similarity loss}.

\textbf{Consistent Similarity Loss} It is expected that the learned codes should follow the smoothness assumption in that a noisy input query correspond to the consistent retrieval results. To what follows, the similarities of codes between the noisy data pairs should be consistent. As illustrated in Figure \ref{fig:framework}(b.2), if $\mathbf{x}_1, \mathbf{x}_2$ is quite similar and so as $\mathbf{x}_3, \mathbf{x}_4$, the difference between $\mathrm{sim}(H(\mathbf{x}_1), H(\mathbf{x}_3))$ and $\mathrm{sim}(H_T(\mathbf{x}_2), H_T(\mathbf{x}_4))$ should be small. Thus the {\em consistent similarity loss} is defined with
\begin{equation}
l_c (u, u_T) = (u - u_T)^2
\label{eq:ssl_u_p}
\end{equation}
where $l_c(u, u_T)$ are the same as Eq. (\ref{eq:ssl_u}). We rename the $\mathcal{R}_{u}$ as $\mathcal{R}_{up}$ if Eq. (\ref{eq:ssl_u_p}) is introduced.

\textbf{Quantized Similarity Loss} The consistent similarity loss is only able to capture the locally structure of a certain data pair, ignoring the global neighborhood structure between samples. Inspired by the quantization methods in which large amount of information can be compressed with quantization~\cite{jegou2011product}, we quantize the pairwise similarity produced by the ensembled teacher to guide the hash learning. As the quantization procedure is based on global unlabeled data pairs, it is expected that the quantized similarities contain global pairwise information, leading to better learned codes.

We denote $\mathbf{W} \in \{0,1\}^{n \times n}$ as the quantized similarity matrix to be learned, where $n$ is the number of training samples. Denote $W_{ij}$ as the element at $i$th row and $j$th column, thus $W_{ij} = 1$ indicates $\mathbf{x}_i$ and $\mathbf{x}_j$ are {\em pseudo similar pair}, and $0$ otherwise. Considering the teacher output $H_T(\mathbf{x})$ is the ensemble of embedded codes of $\mathbf{x}$, thus $H_T(\mathbf{x})$ can be regarded as precise feature embedding of the data point $\mathbf{x}$. To what follows, we use the teacher output to determine the pseudo similar pairs. The similarity matrix is defined according to the distances of teacher output such that
\begin{equation}
W_{ij} = 
\begin{cases}
1 & u_{Tij} \ge thr \\
0 & u_{Tij} < thr \\
\end{cases}
\label{eq:tg}
\end{equation}
where $u_{Tij} = \mathrm{sim}(H_T(\tilde{\mathbf{x}_i}^{(2)}), H_T(\tilde{\mathbf{x}_j}^{(2)}))$ is defined the same as Eq. (\ref{eq:ssl_u}), $thr$ is the threshold, which is set according to the dataset. In practical applications, the distribution between labeled and unlabeled pairs are expected to be the same. We can set $thr$ such that the ratio of pseudo similar pairs is the same as the ratio of similar pairs among labeled pairs, so that the unlabeled similar pairs generated by the teacher can be almost positive and the distribution of similar pairs are expected to the same as the ground-truth similar pairs.

Given the generated pseudo similarity pairs, we can simply train the student with the pairwise loss shown in Eq. (\ref{eq:pls}) to capture the global structure of the embedded codes in the hamming space. We propose the {\em quantized similarity loss} by defining $l_c$ such that:
\begin{equation}
l_c(u_{12}, u_{T12}) = l(u_{12}, W_{12})
\label{eq:ssl_u_g}
\end{equation}
where $l(\cdot,\cdot)$ has the same form as that defined in Eq. (\ref{eq:pls}). It should be noticed that Eq. (\ref{eq:ssl_u_g}) can be regarded as the ranking loss for the global data pairs to some extent in that similar pairs produced by the teacher are more likely to be pseudo similar pairs, thus they are expected to achieve similar hamming distances during training. We rename $\mathcal{R}_u$ as $\mathcal{R}_{uq}$ if Eq. (\ref{eq:ssl_u_g}) is introduced.

\textbf{Overall Training Loss} The overall training loss is defined the same as Eq. (\ref{eq:ssh}), where $\mathcal{L}_s$ is defined in Eq. (\ref{eq:pls}), and $\mathcal{R}_u$ can be regarded as the combination of {\em consistent similarity loss} and {\em quantized similarity loss} such that
\begin{equation}
\mathcal{R}_u = \mathcal{R}_{up} + \gamma \mathcal{R}_{uq}
\label{eq:ts3h}
\end{equation}

As the teacher outputs in Eq. (\ref{eq:ts3h}) lead to better abstract representations and can model the pairwise information locally and globally, it is expected that the proposed loss can better meet the smoothness assumptions and achieves better codes. Moreover, the hamming distances is accordant with the similarities on both labeled and unlabeled data. 

\textbf{Implementation and Relaxation} Eq. (\ref{eq:ts3h}) conveys that the both the original and the perturbed samples should be fed into the network. For simplicity, we just regard the perturbed data as input, shown in Figure \ref{fig:framework}. 

It is clear that directly optimizing Eq. (\ref{eq:ssh}) is intractable as the discrete constraints are involved. As used in most deep hashing algorithms~\cite{cao2016deep,zhang2017ssdh,yan2017semi}, the simple and efficient way is removing the $\mathrm{sgn}$ function and adding the quantization loss. We reformulate the relaxed problem as follows
\begin{equation}
\min_F \mathcal{L} = \mathcal{L}_{s}^{(r)} + \omega \mathcal{R}_u^{(r)} + 
\eta \frac{1}{|\mathcal{X}|} \sum_{\mathbf{x} \in \mathcal{X}} \| \mathbf{h} - F(\tilde{\mathbf{x}}^{(1)}) \|_1
\label{eq:r_ts3h}
\end{equation}
where $\mathbf{h} = \mathrm{sgn}(F(\tilde{\mathbf{x}}^{(1)}))$, $\mathcal{L}_{s}^{(r)}, \mathcal{R}_{u}^{(r)}$ is the relaxation of Eq. (\ref{eq:pls},\ref{eq:ts3h}) respectively such that
\begin{equation}
\begin{split}
\mathcal{L}_{s}^{(r)} =& \frac{1}{|\mathcal{S}|} \sum_{(i,j) \in \mathcal{S}} l(u^r_{ij}, s_{ij}) \\
\mathcal{R}_{u}^{(r)} =& \frac{1}{|\mathcal{X}|^2} \sum_{\mathbf{x}_1, \mathbf{x}_2 \in \mathcal{X}} \Big[(u^r_{12}- u^r_{T12})^2 + \gamma l(u^r_{12}, W_{12}) \Big] 
\end{split}
\label{eq:rr_ts3h}
\end{equation}

For $\mathcal{L}_s^{(r)}$, we directly remove the $\mathrm{sgn}$ function to compute $u^r_{ij}$, and $u^r_{ij}$ is defined the same as that in Eq. (\ref{eq:pls}). For $\mathcal{R}_u^{(r)}$, we use $u^r_{12} = \mathrm{sim}(F(\tilde{\mathbf{x}_1}^{(1)}), F(\tilde{\mathbf{x}_2}^{(1)})), u^r_{T12} = \mathrm{sim}(F_T(\tilde{\mathbf{x}_1}^{(2)}), F_T(\tilde{\mathbf{x}_2}^{(2)}))$, and $\mathrm{sim}(\mathrm{s}, \mathrm{t}) = -\Arrowvert \frac{\mathbf{s}}{\| \mathbf{s} \|} - \frac{\mathbf{t}}{\| \mathbf{t} \|} \Arrowvert^2$ where $\| \cdot \|$ is the $L_2$ normalization. The use of $L_2$ normalization is inspired by the original Mean Teacher where the consistent output is the normalized classification probabilities. Moreover, the norm of the hashcodes are the same, thus similar normalized feature embeddings correspond to similar hashcodes.

As a result, the consistent pairwise losses can capture both the local and global neighborhood structure. Moreover, semantic information can be embedded with supervised pairwise loss, and the real-valued space is able to be mapped into hamming space with the quantization loss.

\subsection{Mini-batch Optimization}

\begin{algorithm}[t]
\setlength{\abovecaptionskip}{0pt}
\setlength{\belowcaptionskip}{0pt}
\caption{Mini-batch Training of \ptsh}
\label{alg:ts3h}
\begin{algorithmic}[1]
\REQUIRE Input data $\mathcal{X}$, pairwise labels $\mathcal{S}$, parameters $\omega(t), \eta, \gamma, \alpha$
\FOR {$t$ in num-epochs}
\STATE Determine the unsupervised weight $\omega=\omega(t)$
\FOR {each mini-batch $B$}
\FOR {$\mathbf{x}_i \in B$}
\STATE Sample two random perturbations $\tilde{\mathbf{x}_i}^{(1)},\tilde{\mathbf{x}_i}^{(2)}$ 
\ENDFOR
\FOR {$(\mathbf{x}_i, \mathbf{x}_j) \in B\times B$}
\STATE Compute $W_{ij}$ by Eq. (\ref{eq:tg})
\ENDFOR
\STATE Compute mini-batch version of $\mathcal{L}$ such that replacing $\mathcal{X}$ with $B$ in Eq. (\ref{eq:rr_ts3h})
\STATE Update $\theta$ with optimizers, e.g. SGD
\STATE Update $\theta_T$ with Eq. (\ref{eq:t_ema})
\ENDFOR
\ENDFOR
\RETURN learned student and teacher networks
\end{algorithmic}
\end{algorithm}

The training procedure is roughly the same as ~\cite{tarvainen2017mean}. The teacher is the average embedding of the student network and is updated by Eq. (\ref{eq:t_ema}) each iteration, and the student is trained with back-propagation. We use the ramp-up procedure for both the learning rate and the regularization term $\omega = \omega(t)$ in the beginning of training. The training algorithm is summarized in Algorithm \ref{alg:ts3h}.

We mainly focus on training the student network. It is clear that the student can be trained by optimizing Eq. (\ref{eq:r_ts3h}) with SGD. We follow the common practice in which we randomly sample mini-batch to estimate the losses for each iteration. For a mini-batch $B$, we just compute the pairwise losses $\mathcal{L}_{s}^{(r)}, \mathcal{R}_{u}^{(r)}$ within the mini-batch, and so as computing the pesudo similar pairs. It is clear that the complexity of the loss just $O(|B|^2)$, thus the computational cost is not large compared with the computational cost of deep networks. To utilize both the labeled and unlabeled data, the ratio of number of labeled data pairs and unlabeled ones is constant in a mini-batch.

\begin{table*}[t]
\setlength{\abovecaptionskip}{0pt}
\setlength{\belowcaptionskip}{0pt}
\centering
\scriptsize
\begin{tabular}{c|c|cccc|cccc|cccc}
\hline
\multirow{2}{*}{Method} & \multirow{2}{*}{Net} & \multicolumn{4}{c|}{CIFAR-10} & \multicolumn{4}{c|}{Nuswide} & \multicolumn{4}{c}{ImageNet-100$^4$}  \\
\cline{3-14}
                  &                   & 12 bits & 24 bits & 32 bits & 48 bits & 12 bits & 24 bits & 32 bits & 48 bits & 16 bits & 32 bits & 48 bits & 64 bits \\
\hline
\multicolumn{14}{c}{Semi-Supervised Hashing} \\
\hline
 SSDH  & VGG-F & 0.801 & 0.813 & 0.812 & 0.814 & 0.773 & 0.779 & 0.778 & 0.778 & -$^1$ & - & - & - \\
 BGDH  & VGG-F & \textbf{0.805} & 0.824 & 0.826 & 0.833 & \textbf{0.803} & 0.818 & 0.822 & 0.828 & - & - & - & - \\
\hline
 \multirow{2}{*}{\textbf{\ptsh-DSH}}  & \multirow{2}{*}{AlexNet} & 0.798 & \textbf{0.828} & \textbf{0.835} & \textbf{0.843} & 0.752 & 0.774 & 0.783 & 0.789 & \textbf{0.612} & \textbf{0.680} & \textbf{0.697} & \textbf{0.703} \\
 & & \tiny{(+0.056)} & \tiny{(+0.034)} & \tiny{(+0.026)} & \tiny{(+0.023)} & \tiny{(+0.012)} & \tiny{(+0.012)} & \tiny{(+0.019)} & \tiny{(+0.016)} & \tiny{(+0.023)} & \tiny{(+0.032)} & \tiny{(+0.047)} & \tiny{(+0.041)} \\
 \multirow{2}{*}{\textbf{\ptsh-DPSH}} & \multirow{2}{*}{AlexNet} & 0.789 & 0.799 & 0.801 & 0.805 & \textbf{0.803} & \textbf{0.827} & \textbf{0.831} & \textbf{0.842} & 0.397 & 0.542 & 0.618 & 0.634 \\
 & & \tiny{(+0.038)} & \tiny{(+0.028)} & \tiny{(+0.025)} & \tiny{(+0.027)} & \tiny{(+0.004)} & \tiny{(+0.006)} & \tiny{(+0.003)} & \tiny{(+0.009)} & \tiny{(+0.018)} & \tiny{(+0.014)} & \tiny{(+0.027)} & \tiny{(+0.026)} \\
\hline
\multicolumn{14}{c}{Supervised Hashing Baselines} \\
\hline
 DSH$^2$     & AlexNet  & 0.741 & 0.794 & 0.809 & 0.820 & 0.740 & 0.762 & 0.764 & 0.773 & 0.589 & 0.648 & 0.650 & 0.662 \\
 DPSH$^2$    & AlexNet    & 0.751 & 0.771 & 0.776 & 0.778 & 0.799 & 0.821 & 0.827 & 0.834 & 0.379 & 0.528 & 0.591 & 0.608 \\
 DSDH    & VGG-F    & 0.740 & 0.786 & 0.801 & 0.820 & 0.776 & 0.808 & 0.820 & 0.829 & - & - & - & - \\
 DISH    & AlexNet  & 0.758 & 0.784 & 0.799 & 0.791 & 0.787 & 0.810 & 0.810 & 0.813 & - & - & - & - \\
 HashNet$^3$ & AlexNet  & 0.686${^3}$ & - & 0.692 & 0.718 & 0.733$^3$ & - & 0.755 & 0.762 & 0.502 & 0.622 & 0.661 & 0.682 \\
 DMDH$^3$    & AlexNet  & 0.704${^3}$ & - & 0.732 & 0.737 & 0.751$^3$ & - & 0.781 & 0.789 & 0.513 & 0.612 & 0.673 & 0.692 \\
 MIHash  & AlexNet  & 0.738 & 0.775 & 0.791 & 0.816 & 0.773 & 0.820 & 0.831 & 0.843 & 0.569 & 0.661 & 0.685 & 0.694 \\
\hline
\end{tabular}
\begin{tablenotes}
\item 1: Results not available. 2: Our own implementation on the three datasets and most results are better than previously reported. 3: Results of HashNet, DMDH are referenced from DMDH~\cite{chena2018deep}. 4: Results at 16 bits.
\end{tablenotes}
\caption{Accuracy in terms of MAP for the semi-supervised and supervised hashing methods. The numbers in blankets are the relative gain compared with the baselines. Unless specified, the results are directly drawn from the original papers.}
\label{tab:deephash}
\end{table*}

\section{Experiments}

In this section, we conduct various large-scale retrieval experiments to show the efficiency of the proposed \ptsh{} method. We compare our \ptsh{} method with recent state-of-the-art semi-supervised deep hashing methods on the retrieval performance. Some ablation studies and sensitivity of parameters are also discussed in this section.

\subsection{Datasets and Evaluation Metrics}

We run large-scale retrieval experiments on three image benchmarks: CIFAR-10\footnote{http://www.cs.toronto.edu/\textasciitilde kriz/cifar.html}, Nuswide\footnote{http://lms.comp.nus.edu.sg/research/NUS-WIDE.htm} and ImageNet-100. CIFAR-10 consists of 60,000 $32 \times 32$ color images from 10 object categories. ImageNet-100 is the subset of ImageNet dataset\footnote{http://image-net.org} with 100 randomly sampled classes. Nuswide dataset contains about 220K available images associating with 81 ground truth concept labels. Following~\cite{liu2011hashing}, we only use the images associated with the 21 most frequent concept tags, where the total number of images is about 190K.

The experimental protocols is similar with~\cite{xia2014supervised}. In CIFAR-10 dataset, we randomly select 1,000 images (100 images per class) as query set, the rest 59,000 images as retrieval database, and we random select 5,000 images from the database as the training data. In Nuswide dataset, we randomly select 2,100 images (100 images per class) as the query set and 10,500 images as the training set. In ImageNet-100 dataset, we use the same data split as HashNet~\cite{cao2017hashnet} such that 130 images per class(totally 13K images) for training, and all images in the selected classes from the validation set are used as queries. The rest unlabeled data in the databest are regarded as the unlabeled dataset.

As we just consider the pairwise similarity for training, the data pairs are constructed among the training data. For CIFAR-10 and ImageNet-100, similar data pairs share the same semantic label. For Nuswide dataset, similar images share at least one semantic label. The rest data pairs(pairs between unlabeled data and all the database) are regarded as the unlabeled pairs.

Our method is implemented with the PyTorch\footnote{http://pytorch.org/} framework. We adopt the pre-trained AlexNet~\cite{krizhevsky2012imagenet} for deep hashing methods but replace the last fully-connected layer. The images are resized to $224 \times 224$ to train the network. For supervised pairwise loss in Eq. ({\ref{eq:pls}}), we mainly use the DSH loss and DPSH loss and name them as \textbf{\ptsh-DSH} and \textbf{\ptsh-DPSH} respectively. SGD with momentum 0.9 is used for optimization, and the initial learning rate of the last layer is $10^{-3} \sim 10^{-2}$ which is ten times larger of the lower layers. The hyper-parameters $\omega, \mu, \alpha$ is different according to datasets, which are selected with the validation set. We first of all randomly select part of training data as validation set to determine the parameters. For CIFAR-10, we use $\{ \omega=0.8, \gamma=0.5, \eta=0.004 \}$ with DSH loss and $\{ \omega=0.02, \gamma=0.5, \eta=0.01 \}$ with DPSH loss; For Nuswide, we use $\{ \omega=0.8, \gamma=0.1, \eta=0.01 \}$ with DSH loss and $\{ \omega=0.2, \gamma=0.1, \eta=0.01 \}$ with DPSH loss. For ImageNet-100, we use $\{ \omega=0.5, \gamma=0.1, \eta=0.004 \}$ with DSH loss and $\{ \omega=0.5, \gamma=0.02, \eta=0.004 \}$ for DPSH loss. Following~\cite{tarvainen2017mean}, we set $\alpha=0.995$, and the ratio of number of unlabeled data pairs and labeled data pairs within a minibatch is 15. The image perturbation strategy includes random resize, random cropping, random horizontal flipping, etc. The training is done on a server with two Intel(R) Xeon(R) E5-2683 v3@2.0GHz CPUs, 256GB RAM and a Geforce GTX TITAN Pascal with 12GB memory. We train 60 epochs for CIFAR-10, 20 epochs for Nuswide, and 240 epochs for ImageNet-100. We apply center cropped input and the teacher network to generate hashcodes for simplicity, and Section \ref{sec:ablation} shows that there are little difference between codes generated by the teacher and the student.

Similar with~\cite{xia2014supervised,cao2017hashnet}, for each retrieval dataset, we report the compared results in terms of {\em mean average precision}(MAP), precision at Hamming distance within 2, precision of top returned candidates. We calculate the MAP value within the top 5000 returned neighbors for NusWide and top 1000 for ImageNet-100, and report the MAP of all retrieved samples on CIFAR-10. Groundtruths are defined by whether two candidates are similar. We run each experiment for 5 times and get the average result.

\begin{figure}[t]
    \setlength{\abovecaptionskip}{0pt}
    \setlength{\belowcaptionskip}{0pt}
    \centering
    \includegraphics[scale=0.225]{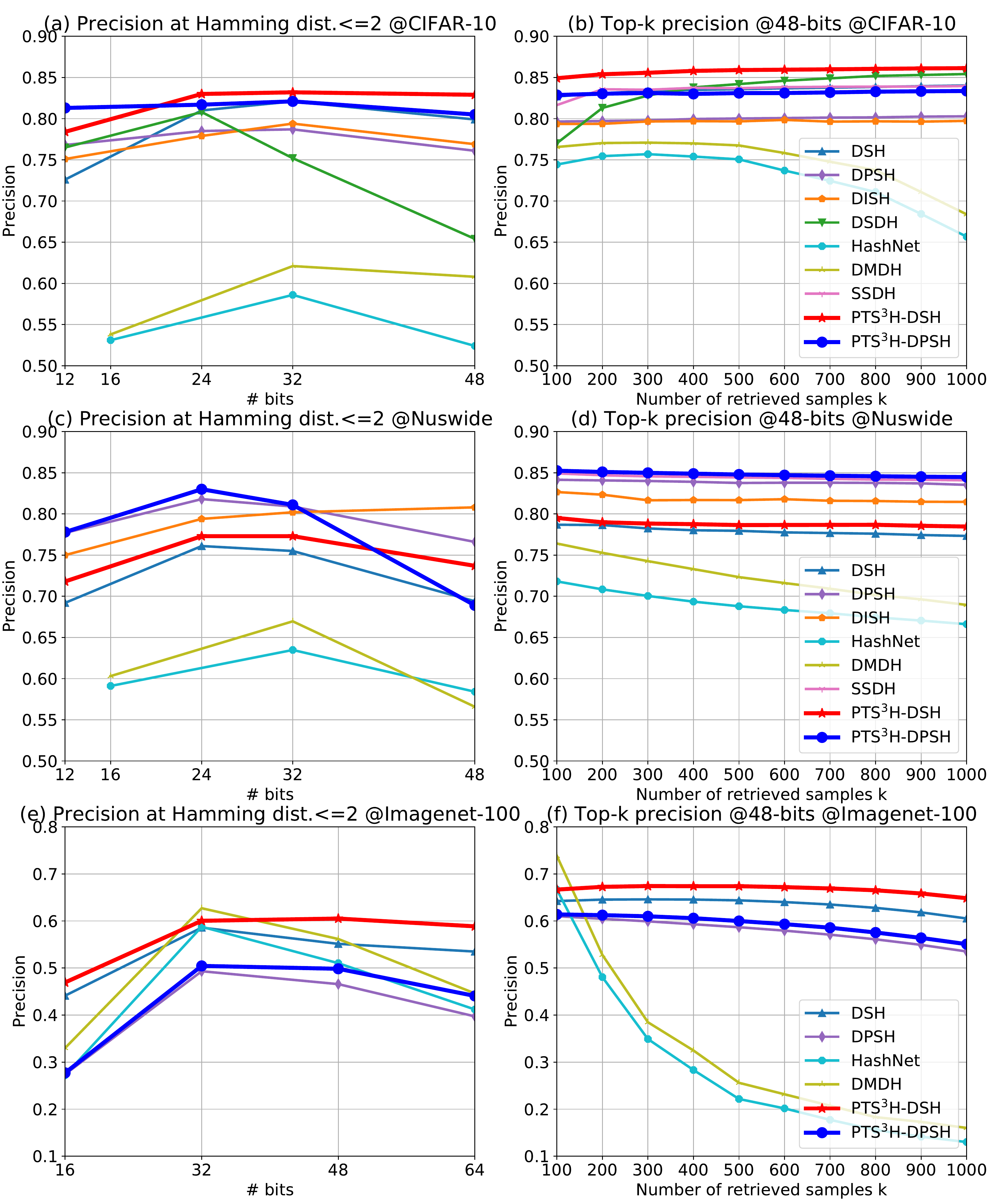}
    \caption{Precision at Hamming distance within 2 value and top-k precision curve of different deep hashing methods. AlexNet/VGG-F is used for pre-training in these algorithms.}
    \label{fig:deephash}
\end{figure}

\subsection{Results}

We compare our \ptsh{} method with recent state-of-the-art deep hashing methods including SSDH~\cite{zhang2017ssdh}, BGDH~\cite{yan2017semi}. We do not take DSH-GANs~\cite{qiu2017deep} into consideration as it utilizes the label of each data point. Results on other supervised hashing methods like DSH~\cite{liu2016deep}, DPSH~\cite{li2015feature}, DSDH~\cite{li2017deep}, DISH~\cite{zhang2017scalable}, DMDH~\cite{chena2018deep} and MIHash~\cite{cakir2018hashing} are also proposed for comparison. They follow similar settings, and the network used is either VGG-F or AlexNet, which share similar architectures. Table \ref{tab:deephash} conveys that DSH and DPSH are good supervised hashing algorithms, we therefore regard the DSH and DPSH loss as the baselines of \textbf{\ptsh-DSH} and \textbf{\ptsh-DPSH} respectively. We report the supervised baselines so that the relative gains of the \ptsh{} are also taken into consideration.

Retrieval results of different methods are shown in Table \ref{tab:deephash} and Figure \ref{fig:deephash}. We re-implement the DSH and DPSH algorithms for all the datasets, and most results of the two baselines are better than previously reported. Note that the settings of Imagenet-100 are the same as that in~\cite{cao2017hashnet}. With the network structure and the training loss fixed, the proposed \ptsh{} algorithm performs much better than the baselines by about 1-5 percents on MAP and precision at Hamming distance within 2 value, which conveys that the proposed semi-supervised setting is able to capture more semantic information with unlabeled data. Moreover, our semi-supervised algorithm achieves much better retrieval performance by a large margin at most bits if proper supervised baselines are selected (DSH for CIFAR-10,ImageNet-100 and DPSH for Nuswide), showing the effectiveness of the proposed teacher-student architecture. 

It should be noticed that the classification performance of VGG-F is slightly better than AlexNet, thus the hashing performance is expected not to decrease and may even be better if replacing AlexNet with VGG-F. Moreover, the proposed baselines are widely used but not the state-of-the-art, thus it is expected to achieve better results if adopting the state-of-the-art supervised hashing methods~\cite{chena2018deep}.

\begin{table}[t]
    \setlength{\abovecaptionskip}{0pt}
    \setlength{\belowcaptionskip}{0pt}
    \centering
    \scriptsize
    \begin{tabular}{c|c|cc|cc}
	    \hline
        \multirow{2}{*}{Method} & \multirow{2}{*}{Dataset} & \multicolumn{2}{c}{MAP} & \multicolumn{2}{|c}{Precision} \\
        &  & 32 bits & 48 bits & 32 bits & 48 bits \\
        \hline
        \ptsh-P & \multirow{4}{*}{CIFAR-10} & 0.829 & 0.838 & 0.829 & 0.827 \\
        \ptsh-Q & & 0.817 & 0.826 & 0.821 & 0.814  \\
        \cline{1-1} \cline{3-6}
        \textbf{\ptsh} & & \textbf{0.835} & \textbf{0.843} & 0.832 & 0.829  \\
        \textbf{\ptsh-S} & & 0.833 & 0.842 & \textbf{0.834} & \textbf{0.830} \\
        \hline
        \ptsh-P & \multirow{4}{*}{Nuswide} & 0.777 & 0.787 & 0.763 & 0.727  \\
        \ptsh-Q & & 0.772 & 0.777 & 0.759 & 0.710 \\
        \cline{1-1} \cline{3-6}
        \textbf{\ptsh} & & 0.782 & \textbf{0.789} & 0.770 & 0.737 \\
        \textbf{\ptsh-S} &  & \textbf{0.783} & \textbf{0.789} & \textbf{0.771} & \textbf{0.739} \\
        \hline
        
    \end{tabular}
    \caption{Results of the variants of the proposed \ptsh{} algorithm on CIFAR-10 and Nuswide dataset. \ptsh{} and \ptsh-S are both proposed method but the codes are generated by the teacher and the student respectively. AlexNet is used for pre-training. Precision denotes the precision at Hamming distance within 2 value.}
    \label{tab:variant_deep}
\end{table}

\subsection{Ablation Study}
\label{sec:ablation}

\textbf{Variants of \ptsh} In order to verify the effectiveness of our \ptsh{} method, several variants are also considered. First we set $\gamma=0$ to show the effectiveness of the $\mathcal{R}_{up}$, named \ptsh-P. Then we remove $\mathcal{R}_{up}$ to show the effectiveness of $\mathcal{R}_{uq}$, denote \ptsh-Q. The hyper-parameters of the variants are determined with the validation set. Retrieval results are shown in Table \ref{tab:variant_deep}. The {\em consistent similarity loss} reaches about 70\% performance gain as it produces consistent simialrities for smooth data pairs. The {\em quantized similarity loss} also achieves better performance as they model the pairwise similarities for perturbed inputs with global information. It should be noticed that there are little performance gain on MAP with the {\em quantized similarity loss} for Nuswide dataset, as the distribution of similar pairs underlying the dataset is a little complicated. Better results may achieved if better similarity construction strategy is involved.

\textbf{The Teacher vs. the Student} We denote \ptsh{} and \ptsh-S as hashcodes generated by the teacher (denote $H_T(\cdot)$) and the student (denote $H(\cdot)$) respectively. Table \ref{tab:variant_deep} shows retrieval results of \ptsh{} and \ptsh-S. It implies that the performances are almost the same, thus we are able to use the teacher or the student freely. As the student is converged during training, the teacher will be similar with the student in the end of training. Nevertheless, the parameters of the teacher and the student are quite different during training. As the teacher is the ensemble of the student, the representations generated by the teacher are expected to contain more semantic information than the student at the most training stage~\cite{tarvainen2017mean}, guiding the student to generate better codes.

\begin{figure}[t]
    \setlength{\abovecaptionskip}{0pt}
    \setlength{\belowcaptionskip}{0pt}
    \centering
    \includegraphics[scale=0.23]{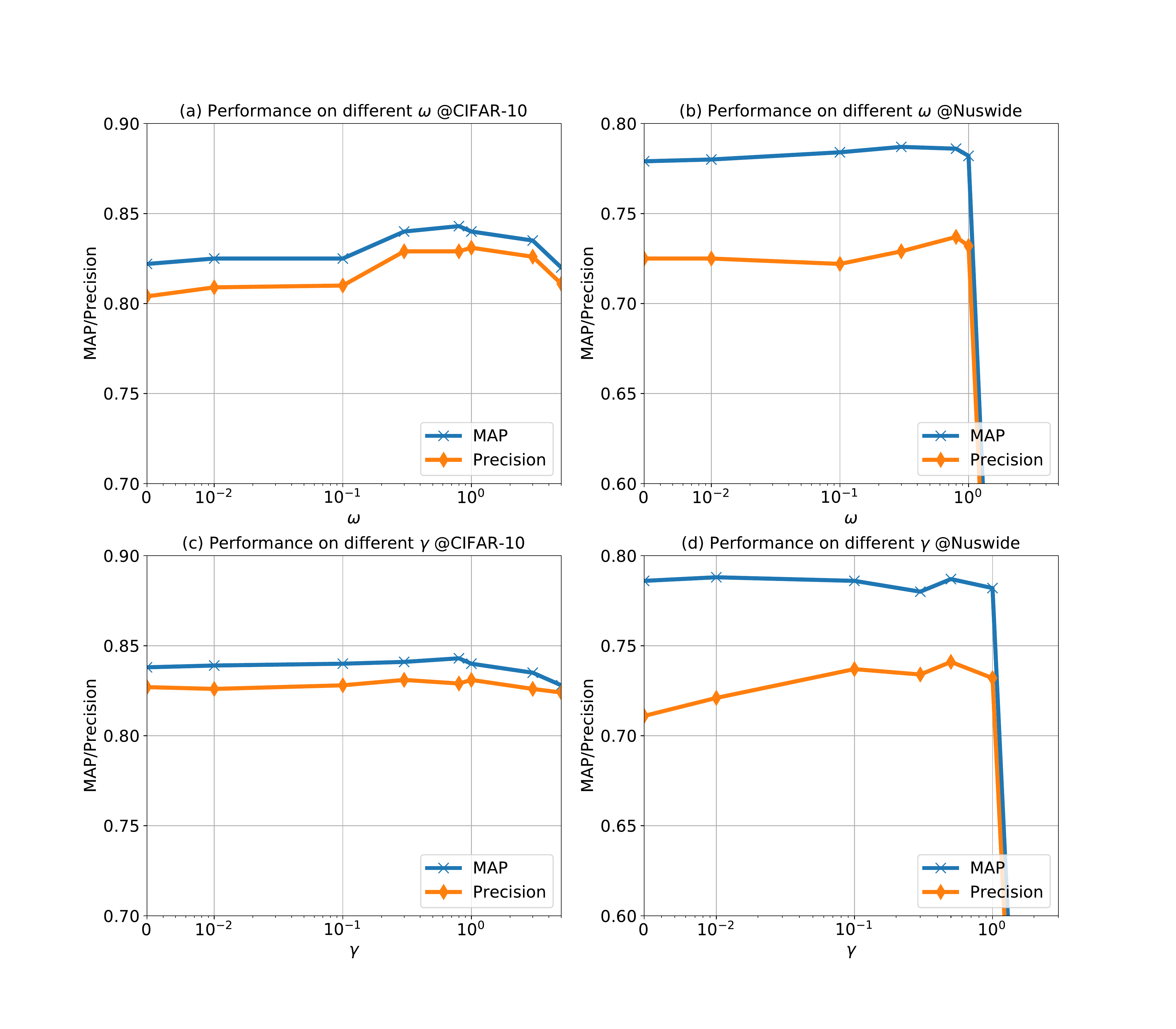}
    \caption{Comparative results of different hyper-parameters on CIFAR-10 and Nuswide dataset. The code length is 48. We use the DSH loss for training.}
    \label{fig:param}
\end{figure}

\begin{figure}[t]
    \setlength{\abovecaptionskip}{0pt}
    \setlength{\belowcaptionskip}{0pt}
    \centering
    \includegraphics[scale=0.23]{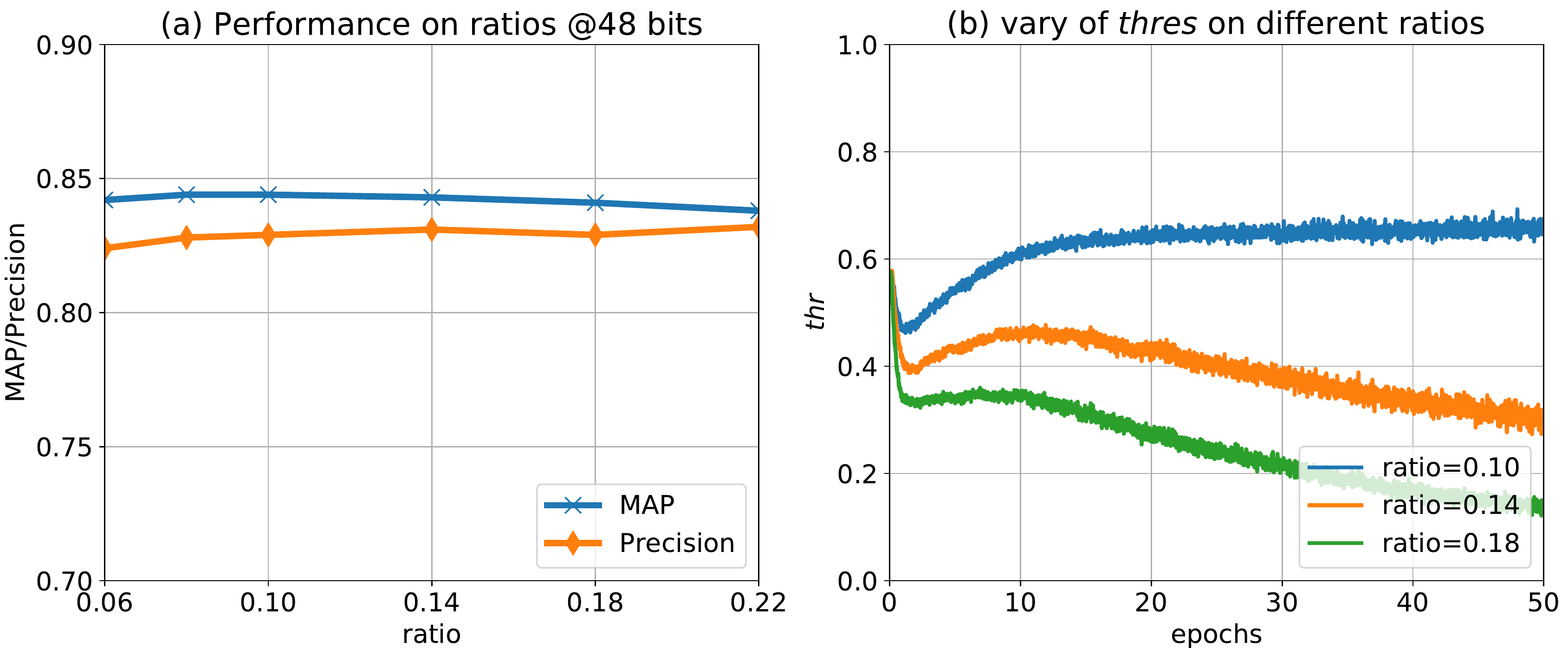}
    \caption{Comparative results of different ratio of pseudo similar pairs in the unlabeled pairs and the corresponding variation of $thr$. We use the DSH loss for training CIFAR-10 and the code length is 48.}
    \label{fig:graph_thr}
\end{figure}

\subsection{Sensitivity to Parameters}

In this section, the influence on different setting of the proposed \ptsh{} is evaluated. The code length is 48 and we use DSH loss for evaluation. We do not report the influence on $\eta$ as it has been discussed in the original papers~\cite{liu2016deep,li2015feature}.

\textbf{Influence of $\omega$} Figure \ref{fig:param}(a)(b) shows the performance on different values of $\omega$. It can be seen clearly that setting a certain $\omega$ achieves better hashing performance. It means that a proper consistent weight $\omega$ can arrive at better semi-supervised training.

\textbf{Influence of $\gamma$} Figure \ref{fig:param}(c)(d) shows the performance on different values of $\gamma$. It should be noticed that a proper $\omega$ is set for different $\gamma$. There are some improvement for a proper $\gamma$, especially the precision at Hamming distance within 2 value on Nuswide dataset. Similar as $\omega$, a proper $\gamma$ should be set for better performance.

\textbf{Influence of $thr$} As discussed in Sec. \ref{sec:loss}, the $thr$ is set dynamically such that the ratio of pseudo similar pairs of unlabeled data is constant. Figure \ref{fig:graph_thr} shows the performance on different ratio value and the variation of $thr$ during training. It is clear that performance is not sensitive for different ratio of pseudo similar pairs $thr$, thus we can set this parameter freely.

\section{Conclusion and Future Work}

In this paper, we propose a novel semi-supervised hashing algorithm named \ptsh{} in which the pairwise supervision and abundant unlabeled data are provided. The proposed \ptsh{} is a teacher-student network architecture which is carefully designed for labeled and unlabeled pairs. We propose the general {\em consistent pairwise loss} in which the pairwise information generated by the teacher network guides the training of the student. There are two types of losses: {\em consistent similarity loss} models the locally pairwise information, and {\em quantized similarity loss} models the information globally by quantizing the similarities between samples. This procedure aims at generating similar retrieval results for neighborhood queries. Experiment shows that the proposed \ptsh{} achieves great improvement over the baselines, and it is superior or comparable with the state-of-the-art semi-supervised hashing algorithms.

It should be noticed that we use the popular pairwise loss baselines and achieve the good hashing results. As the proposed \ptsh{} algorithm is a general framework for semi-supervised hashing, it is expected to arrive at better retrieval performance by incorporating the state-of-the-art supervised hashing algorithm with pairwise supervisions.

{\small
\bibliographystyle{ieee}
\bibliography{references}
}

\end{document}